\pdfoutput=1

\documentclass[11pt]{article}

\usepackage[]{emnlp2021}

\usepackage{times}
\usepackage{latexsym}

\usepackage[T1]{fontenc}

\usepackage[utf8]{inputenc}

\usepackage{microtype}

\usepackage{booktabs}
\usepackage{multirow}
\usepackage{pifont}
\usepackage{xcolor}
\usepackage{xspace}
\usepackage{makecell}
\usepackage{graphicx}

\newcommand{\YES}{\ding{51}}
\newcommand{\NO}{{\color{black!35}\ding{55}}}

\mathchardef\mhyphen="2D
\def\MultiLexNorm{W-NUT 2021:\ Multilingual Lexical Normalization (MultiLexNorm)\xspace}
\newenvironment{citemize}{\begin{list}{$\bullet$}{\topsep=.2\smallskipamount\itemsep=0pt\parsep=1pt\labelwidth=.5em}}{\end{list}}
\newenvironment{cenumerate}{\begin{list}{\labelenumi}{\usecounter{enumi}\topsep=.2\smallskipamount\itemsep=0pt\parsep=1pt\labelwidth=1.0em}}{\end{list}}

\title{ÚFAL at MultiLexNorm 2021: Improving Multilingual Lexical Normalization by Fine-tuning ByT5}

\author{
  David Samuel \and Milan Straka\\
  Charles University, \\
  Faculty of Mathematics and Physics, \\
  Institute of Formal and Applied Linguistics \\
  \texttt{\{samuel,straka\}@ufal.mff.cuni.cz}
}

\usepackage[absolute]{textpos}
\begin{document}
\begin{textblock}{16}(0,0.1)\centerline{This paper was published in \textbf{W-NUT 2021} -- please cite the published version {\small\url{https://aclanthology.org/2021.wnut-1.54}}.}\end{textblock}
\maketitle
\begin{abstract}

We present the winning entry to the \textit{Multilingual Lexical Normalization (MultiLexNorm)} shared task at \textit{W-NUT 2021} \cite{multilexnorm}, which
evaluates lexical-normalization systems on 12 social media datasets in 11
languages. 
We base our solution on a pre-trained byte-level language model, ByT5~\cite{xue-etal-2021-byt5}, which we further pre-train on synthetic data and then fine-tune on authentic normalization data. Our system achieves the best
performance by a wide margin in intrinsic evaluation, and also the best
performance in extrinsic evaluation through dependency parsing.
The source code is released at
{\small\url{https://github.com/ufal/multilexnorm2021}} and the fine-tuned
models at {\small\url{https://huggingface.co/ufal}}.

\end{abstract}

\section{Introduction}

People produce text in natural language every minute of every day. However, in
many cases, for example on social media like Twitter, such texts are not
conforming to a formal style. Instead, they are in colloquial form, which is
perfectly understandable to other people, but challenging for automatic
natural language processing. To make the processing of such texts more viable,
the task of \textit{lexical normalization} can be used to replace the input
forms with their canonical (more formal, lexically normalized) variants.

The aim of the \textit{\MultiLexNorm} shared task \cite{multilexnorm} is to evaluate
participant lexical-normalization systems on 12 social media datasets in 11
languages, including two code-switching datasets. Both intrinsic and extrinsic
evaluation is performed, where the latter is measured through dependency
parsing performed on the normalized data.

Recently, large pre-trained models like BERT~\cite{devlin-etal-2019-bert} or
T5~\cite{raffel-etal-2020-exploring} have demonstrated superior performance in
many NLP tasks when trained in a transfer learning setting. In line with that, we approach the lexical normalization shared task as a fine-tuning of a large pre-trained model, specifically the multilingual byte-level generative language model ByT5~\cite{xue-etal-2021-byt5}.

Our system achieves the best performance in the shared task, both in
intrinsic and extrinsic evaluations. In the intrinsic evaluation, our system
reaches 67.3\% error reduction compared to leave-as-is baseline; the
second-best system has 53.6\%.

The source code is released at {\small\url{https://github.com/ufal/multilexnorm2021}}
and the fine-tuned models 
are available in HuggingFace
Transformers~\cite{wolf-etal-2020-transformers} at
{\small\url{https://huggingface.co/ufal}}.

\begin{figure*}[t]
  \centering
  \includegraphics[width=\hsize]{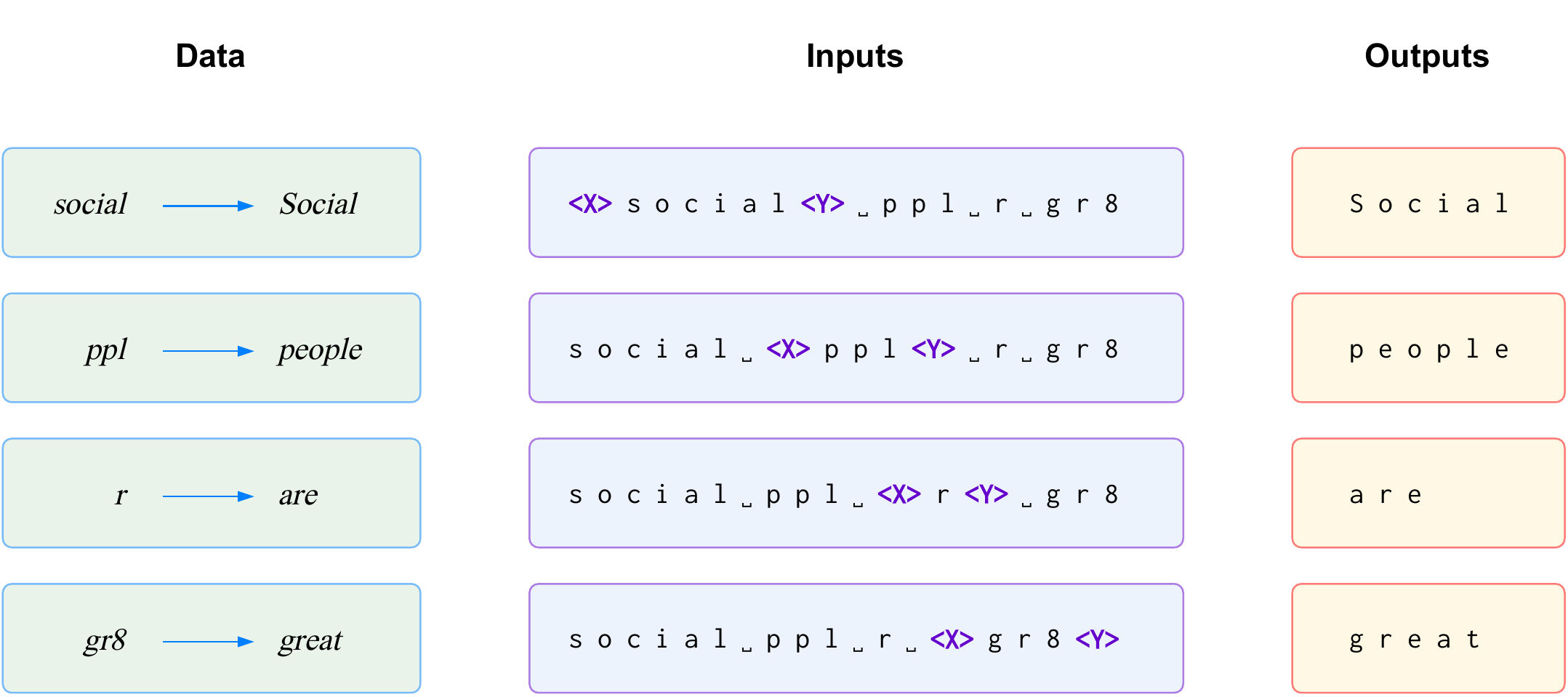}
  \caption{The inputs and outputs of our lexical normalization model. The ByT5 sentinel tokens \texttt{<X>} and \texttt{<Y>} mark the word to be normalized.}
  \label{fig:byt5-multilexnorm}
\end{figure*}

\section{Related Work}

Lexical normalization can be considered a simplified variant of a well-studied
problem of \textit{grammar error correction} (GEC).
\citet{grundkiewicz-etal-2019-neural} approach GEC as a neural machine
translation task using the Transformer
architecture~\citep{vaswani-etal-2017-attention}, which is pre-trained using
a vast amount of synthetic data generated by character-level and word-level
edits. Recently, \citet{rothe-etal-2021-simple} presented a GEC system
based on multilingual mT5~\cite{xue-etal-2021-mt5}, reaching state-of-the-art
results on several datasets with the gigantic \textit{xxl} model size with 13B
parameters.

While the mentioned GEC systems are autoregressive, lexical normalization can
be easily solved in a non-autoregressive way, because the normalizations
of different words are relatively independent. Besides, successful
non-autoregressive models have been recently proposed for general GEC
\citep{awasthi-etal-2019-parallel,omelianchuk-etal-2020-gector}.

Although fine-tuned language models have been successfully used in the state-of-the-art GEC systems, this has not been the case in the field of lexical normalization \citep{muller-etal-2019-enhancing, lourentzou2019adapting}. MoNoise~\cite{van-der-goot-2019-monoise} is a publicly available multilingual
lexical normalization tool achieving competent performance (an improved version
of the system would place third in the shared task intrinsic evaluation).
It utilizes Aspell 
dictionaries, FastText~\cite{bojanowski-etal-2017-enriching} embeddings and hand-crafted language features, but no
contextualized embeddings.

\section{Model}

Our model is based on a large pre-trained multilingual model. In accordance
with \citet{bommasani-etal-2021-on}, we call such models (like BERT or T5) the
\textit{foundation} models.

Specifically, we utilize the ByT5~\cite{xue-etal-2021-byt5} foundation model.
It is a byte-level generative sequence-to-sequence model, which processes
a sequence of bytes of UTF-8 encoding on input and produces a sequence of UTF-8
encoding bytes on output. ByT5 models were proposed as an alternative to multilingual
mT5 foundation models~\cite{xue-etal-2021-mt5} and have been shown to perform
remarkably well on noisy text data (compared to subword-based mT5), like
TWEETQA~\cite{xiong-etal-2019-tweetqa}.

\subsection{Input and Output Format}

\begin{figure}[t]
  \centering
  \includegraphics[width=\hsize]{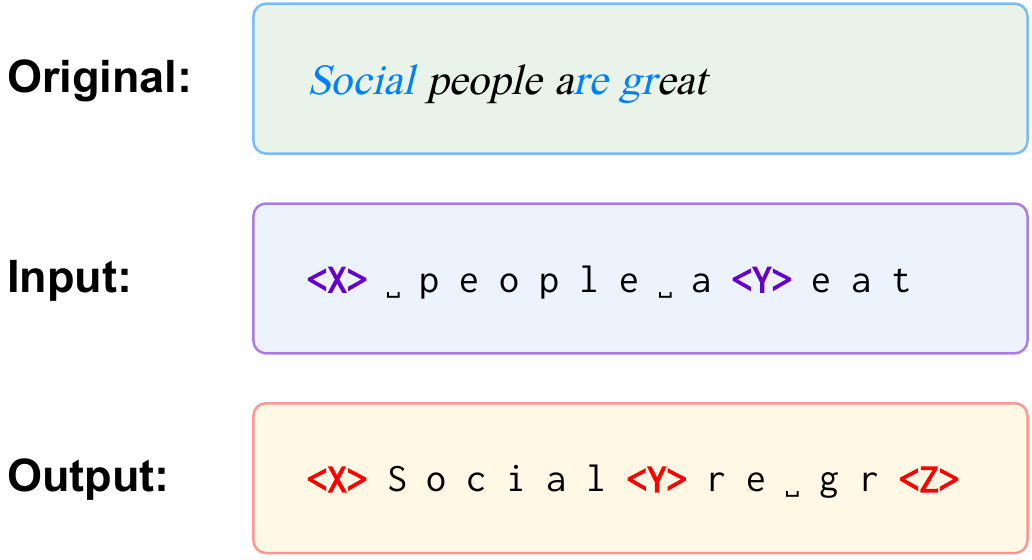}
  \caption{The pre-training task utilized in the ByT5 model~\cite{xue-etal-2021-byt5}.
  The \texttt{<X>}, \texttt{<Y>} and \texttt{<Z>} are the sentinel tokens.}
  \label{fig:byt5}
\end{figure}

We start by recapitulating the pre-training task of the ByT5 model. The input
sentence (including spaces) is represented as a sequence of bytes of UTF-8
encoding, and spans of around 20 bytes are masked using special
\textit{sentinel tokens}. The goal of the model is to reconstruct all the
masked spans. We illustrate the task visually in Figure~\ref{fig:byt5}.

For lexical normalization, we could directly use the unnormalized sentence
as input and the normalized sentence as output (an approach used by
\citet{rothe-etal-2021-simple} for GEC). However, we were concerned that such
an approach would be too different from the ByT5 pre-training, and furthermore,
it would not allow to reconstruct the alignment of the normalized tokens
when a word is removed during normalization or split into several words.

Instead, we opted for a different approach illustrated in
Figure~\ref{fig:byt5-multilexnorm}. For each input word, we construct
a separate ByT5 input, in which we mark the beginning and the end of the word
in question using two sentinel tokens. Then we ask the model to produce just the
normalization of that word. Such an approach normalizes each input word
independently, and we consider it quite similar to the original ByT5
pre-training task. Unfortunately, it requires to encode different input
sequences for every input word, which is considerably inefficient, even if we
can perform normalization of all words in parallel.

\subsection{Pre-training on Synthetic Data}
\label{sec:synthetic-data}

Fine-tuning a ByT5 model directly with supervised training data would not
deliver very high performance, given that the normalization task is (despite our
efforts) still quite different from the pre-training task, and that the amount of
available training data is quite low.\footnote{This experiment is included
in ablations in Section~\ref{sec:ablations}.}

\begin{figure}[t]
  \centering
  \includegraphics[width=\hsize]{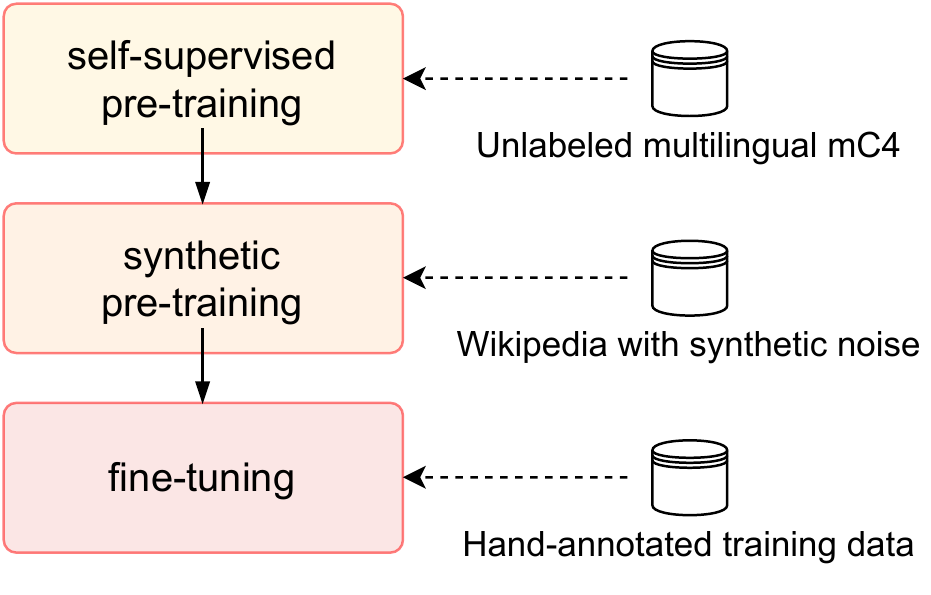}
  \caption{Three stages of training of our models: the original ByT5 pre-training, pre-training on synthetic data and final fine-tuning of authentic data.}
  \label{fig:stages}
\end{figure}

\looseness-1
Therefore, before fine-tuning, we first pre-train the ByT5 model using synthetic data, as illustrated in Figure~\ref{fig:stages}. Note that
from now on, by \textit{``pre-train''} we mean the training of the ByT5 foundation
model using synthetic lexical normalization data.

We construct the synthetic data by modifying clean data in the following way:
\begin{citemize}
  \item If a word is present in the normalized output in the training data, it
    is replaced by one of the corresponding (possibly unnormalized) training data
    inputs, proportionally to the number of occurrences. For example, we change \textit{``people''} to \textit{``ppl''} with 39.7\% probability in the English dataset or \textit{``ikke''} to \textit{``ik''} with 4.97\% chance in the Danish dataset.
  \item A large portion of the required changes can be realized as simple character-level modifications. Thus, we synthesize the data by reversing these alterations. These are 1) accent mark removal (e.g., replacing \textit{``š''} by \textit{``s''} with 16.3\% chance in Slovenian), 2) changing capitalization (e.g., lowering the first capitalized letter with 1.0\% chance in Turkish), 3) removing apostrophes (46.8\% probability in English), and 4) other  miscellaneous modifications (e.g., simplifying \textit{``qu''} to \textit{``k''} in Spanish with 2.0\% probability).
  \item The training data naturally contain a large quantity of different typographical errors, usually caused by inaccurate typing on a keyboard. To simulate this behavior, we alter the words by skipping some characters, changing or inserting some characters (more likely to those close on a computer keyboard), and by reversing two consecutive characters. To give one example, the probability of making an artificial typo in Italian is 0.458\%.
  \item Some modifications are unique for a specific language. For example, the plural forms in Indonesian can be created by duplicating the singular forms. As users want to save time when writing in Indonesian, they sometimes indicate the plural by simple appending ``2'' to the singular form. We reverse this transformation with 33.3\% probability (thus, \textit{``laki-lakinya''} becomes \textit{``laki2nya''}).
  \item 7 datasets in this shared task split or merge words. We synthetically split/merge words on the clean dataset to model these actions. For example, the probability of merging two words in Dutch is 5.99\% and of splitting a word is 0.0565\%. 
  \item In order to speed up typing, the users modify their language in various ways -- e.g. by omitting vowels or shortening words to prefixes. On the other hand, they repeat certain characters to make their messages more expressive. We include these variations in the synthetic datasets to get closer to the real data.
\end{citemize}
\noindent For more details, please consult the source code.

\begin{table*}
  \centering
  \setlength{\tabcolsep}{4.5pt}
  \resizebox{\textwidth}{!}{%
  \begin{tabular}{@{}lllrccrr@{}}\toprule
    \textbf{Code} & \textbf{Language} & \textbf{Original Source} & \textbf{Words} & \small\makecell[c]{\textbf{Words}\\\textbf{split/merged}} & \textbf{Caps} & \small\makecell[c]{\textbf{\% words}\\\textbf{normed}} & \small~\makecell[c]{\textbf{MFR}\\\textbf{ERR}}~ \\
  \midrule
    DA & Danish             & \citet{plank-etal-2020-dan} & 11~816 & \YES & \YES & 9.25  & 49.68 \\
    DE & German             & \citet{sidarenka2013rule} & 25~157 & \YES & \YES & 17.96  & 32.09 \\
    EN & English            & \citet{baldwin-etal-2015-shared} & 73~806 & \YES & \NO  & 6.90  & 64.93 \\
    ES & Spanish            & \citet{alegria2013introduccion} & 13~827 & \NO  & \NO  & 7.69  & 25.57 \\
    HR & Croatian           & \citet{11356/1170} & 75~276 & \NO  & \NO & 8.89  & 36.52 \\
    ID-EN & Indonesian-English & \citet{barik-etal-2019-normalization} & 23~124 & \YES & \NO  & 12.16 & 61.17 \\
    IT & Italian            & \citet{van-der-goot-etal-2020-norm} & 14~641 & \YES & \YES & 7.32  & 16.83 \\
    NL & Dutch              & \citet{dutchNorm} & 23~053 & \YES & \YES & 28.29 & 37.70 \\
    SR & Serbian            & \citet{11356/1171} & 91~738 & \NO  & \NO & 7.65  & 42.62 \\
    SL & Slovenian          & \citet{11356/1123} & 75~276 & \NO  & \NO & 15.62 & 56.71 \\
    TR & Turkish            & \citet{colakoglu-etal-2019-normalizing} & 7~949  & \YES  & \YES & 37.02 & 14.53 \\
    TR-DE & Turkish-German     & \citet{van-der-goot-etal-2021-lexical} & 16~546 & \YES & \YES & 24.14 & 22.09 \\
  \bottomrule
  \end{tabular}%
  }
  \caption{The MultiLexNorm datasets and their properties -- number of words, whether words are split/merged, if capitalization is corrected, relative number of normalized words and performance of the most-frequent-replacement baseline (Section~\ref{sec:results}).}
  \label{tbl:data}
\end{table*}

It is important to note that all the probabilities were estimated from the training data. Therefore the synthetic pre-training cannot be regarded as completely unsupervised; one would need expert knowledge to extend this type of pre-training on a language without annotated data of lexical normalization.

We need to make sure that most of the original data (before the synthetic modifications) are clean. Since we also want an ample amount of cross-lingual text, we opted to use publicly available dumps of Wikipedia.\footnote{\scriptsize\url{https://dumps.wikimedia.org/backup-index.html}} We remove any lines shorter than 32 characters or ending with a colon (to remove titles), segment the lines into sentences with Stanza \cite{qi2020stanza} and tokenize with \texttt{CMU-ARK tokenizer}\footnote{\scriptsize\url{https://github.com/myleott/ark-twokenize-py}} (a tool standardly used for tokenizing text from Twitter).


\subsection{Fine-tuning}
\label{sec:fine-tuning}

The final fine-tuning of the pre-trained model can be straightforwardly
performed on the training data (so called \texttt{base} fine-tuning). However, we also consider fine-tuning
on a mix of both authentic training data and the synthetic data (with
1:1 ratio) in order to avoid overfitting (\texttt{mixed} fine-tuning).

\subsection{Inference}

To predict the normalized sentences during inference, each word-token is processed independently, similarly to training (Figure \ref{fig:byt5-multilexnorm}). The ByT5 decoder autoregressively generates each token, either greedily or via beam search.\footnote{For simplicity, we use the default settings of the \texttt{greedy\_search} and \texttt{beam\_search} procedures from the Hugging Face library, apart from allowing longer sequences and varying the number of beams.} When using the beam search, we generate multiple candidate sequences, and the beam search automatically assigns prediction scores to each of them. These can then be used to aggregate predictions from multiple models in an ensemble. 

\section{Experiments}
\label{sec:training}

The MultiLexNorm shared task consists of 12 social media datasets in 11 languages, including two code-switching datasets. All these datasets are based on Twitter data, with the Danish and the Dutch ones also including data from other sources. The datasets, their original sources and some of their properties are listed in Table~\ref{tbl:data}.

We train an independent model for every dataset. The pre-training starts from the \textit{small} variant of ByT5, which contains 300M parameters. We keep all hyperparameters of the model unchanged, including a dropout rate of 10\%. The training employs batch size 128 and the AdaFactor optimizer~\cite{shazeer-stern-2018-adafactor} to decrease memory usage.

The pre-training is performed using the synthetic data described in Section~\ref{sec:synthetic-data}, for at least 100$\,$000 optimizer steps.\footnote{We stop the training only after a full epoch, when at least 100k steps were performed.} We utilize inverse square root decay with peak learning rate of $5 \cdot 10^{-4}$ and 4$\,$000 warm-up steps.

The fine-tuning phase is carried out with a constant learning rate $1 \cdot 10^{-4}$ for 50 epochs. We consider two training data configurations:
\begin{cenumerate}
  \item Using only MultiLexNorm training data, respecting the train/dev split (using 10\% of training data as development set if there is none).
  \item Because the development set, if present, is quite large (usually circa 30\% of the training data), we also consider training on development data. Specifically, we concatenate training and development data, if any, and take only 3\% of the data as development set (just to detect errors, because evaluation on such small set is considerably noisy).
\end{cenumerate}
\noindent Our competition model is trained on the combined dataset (the latter option), and the synthetic data is mixed in (\texttt{mixed} fine-tuning from Section \ref{sec:fine-tuning}) if it improves development performance over the \texttt{base} fine-tuning.\footnote{Ablations of all training data configurations are evaluated in Section~\ref{sec:ablations}.}

\begin{table*}[t!]
  \centering
  \setlength{\tabcolsep}{3.5pt}
  \catcode`@ = 13\def@{\bfseries}
  \catcode`! = 13\def!{\itshape}
  \resizebox{\textwidth}{!}{%
  \begin{tabular}{@{}lrrrrrrrrrrrrr@{}}\toprule
    \textbf{Team} & \multicolumn{1}{c}{\textbf{Average}} & \multicolumn{1}{r}{\textbf{DA}} & \multicolumn{1}{r}{\textbf{DE}} & \multicolumn{1}{r}{\textbf{EN}} & \multicolumn{1}{r}{\textbf{ES}} & \multicolumn{1}{r}{\textbf{HR}} & \multicolumn{1}{r}{\textbf{ID-EN}} & \multicolumn{1}{r}{\textbf{IT}} & \multicolumn{1}{r}{\textbf{NL}} & \multicolumn{1}{r}{\textbf{SL}} & \multicolumn{1}{r}{\textbf{SR}} & \multicolumn{1}{r}{\textbf{TR}} & \multicolumn{1}{r@{}}{\textbf{TR-DE}} \\
  \midrule
\textbf{ÚFAL} (ensemble)  & @67.30 & 68.7  & @66.2 & @75.6 & 59.2  & @67.7 & @67.2 & @47.5  & @63.6 & @80.1 & @74.6 & @68.6 & @68.6 \\
\textbf{ÚFAL} (single)    & 66.21  & @70.2 & 65.7  & 73.8  & 55.9  & 67.3  & 66.2  & 42.6   & 62.7  & 79.8  & 73.5  & @68.6 & 68.2 \\
HEL-LJU & 53.58  & 56.6   & 59.8  & 62.0  & 35.5  & 56.2  & 55.3  & 35.6   & 45.9  & 67.0  & 66.4  & 51.2  & 51.2 \\
!MoNoise  &!49.02  &!51.3  &!47.0  &!74.3  &!45.5  &!52.6  &!59.8  &!21.8   &!49.5  &!61.9  &!59.6  &!28.2  &!36.7 \\
TrinkaAI\textsuperscript{*} & 43.75  & 45.9  & 47.3  & 66.0  & @61.3 & 41.3  & 56.4  & 15.8   & 45.7  & 59.5  & 44.5  & 15.5  & 25.8 \\
thunderml\textsuperscript{*}  & 43.44  & 46.5  & 46.6  & 64.1  & 60.3  & 40.1  & 59.1  & 11.9   & 44.0  & 59.3  & 44.5  & 15.9  & 29.0 \\
team  & 40.70  & 48.1  & 46.1  & 63.7  & 21.0  & 40.4  & 59.3  & 13.9   & 43.7  & 60.6  & 46.1  & 15.9  & 29.7 \\
learnML & 40.30  & 40.5  & 43.7  & 61.6  & 56.5  & 38.1  & 56.2  & 5.9    & 42.8  & 58.2  & 40.0  & 14.4  & 25.7 \\
maet    & 40.05  & 48.1  & 46.1  & 63.9  & 21.0  & 40.4  & 59.3  & 5.9    & 43.7  & 60.6  & 46.1  & 15.9  & 29.7 \\
!MFR     &!38.37  &!49.7  &!32.1  &!64.9  &!25.6  &!36.5  &!61.2  &!16.8   &!37.7  &!56.7  &!42.6  &!14.5  &!22.1 \\
CL-MoNoise & 12.05  & 7.3   & 16.5  & 4.1   & 5.0   & 26.4  & 2.4   & 0.0    & 16.2  & 8.8   & 20.1  & 17.6  & 20.2 \\
BLUE & 6.73   & 49.7  & -1.9  & 26.8  & -9.4  & -10.1 & -7.2  & -31.7  & -2.1  & -1.0  & 42.6  & 10.0  & 15.0 \\
!LAI      &!0.00   &!0.0   &!0.0   &!0.0   &!0.0   &!0.0   &!0.0   &!0.0    &!0.0   &!0.0   &!0.0   &!0.0   &!0.0 \\
MaChAmp  & -21.25 & -88.9 & -93.4 & 51.0  & 25.4  & 42.6  & 39.5  & -312.9 & 1.5   & 56.8  & 39.4  & -12.7 & -3.4 \\
  \bottomrule
  \end{tabular}%
  }
  \caption{The results of MultiLexNorm intrinsic evaluation. Each team could submit two systems, we show the best of the two. More detailed results are available in the overview paper \citep{multilexnorm} and in the participant papers: HEL-LJU \citep{wnut-sinai}, TrinkaAI \citep{wnut-seqlab}, CL-MoNoise \citep{wnut-clmonoise} and BLUE \citep{wnut-seqseq}. \textsuperscript{*} denotes late submissions.}
  \label{tbl:intrinsic}
\end{table*}

\section{Results}
\label{sec:results}

The MultiLexNorm participants were allowed to submit two runs. The first run of our team \textit{ÚFAL} is a single model, while the second run is an ensemble of 4 models.\footnote{The ensemble models start from a single pre-trained checkpoint and only the fine-tuning is independent.} We perform ensembling by considering for each word and each model 16 replacements including their probabilities (using a beam-search decoder), and producing the replacement with the highest average probability.

Apart from the participant systems, several baselines are also evaluated: \textit{LAI} (leave-as-is), \textit{MFR} (most frequent replacement based on the training data) and the \textit{MoNoise} tool~\cite{van-der-goot-2019-monoise}.

\begin{table*}[t!]
  \centering
  \catcode`@ = 13\def@{\bfseries}
  \catcode`! = 13\def!{\itshape}
  \resizebox{\textwidth}{!}{%
  \renewcommand{\arraystretch}{0.97}
  \begin{tabular}{@{}lcccccccc@{}}\toprule
    @Team & \multicolumn{1}{c}{@Average} & \multicolumn{1}{c}{\makecell[c]{@DE\\@tweede}} & \multicolumn{1}{c}{\makecell[c]{@EN\\@aae}} & \multicolumn{1}{c}{\makecell[c]{@EN\\@monoise}} & \multicolumn{1}{c}{\makecell[c]{@EN\\@tweebank2}} & \multicolumn{1}{c}{\makecell[c]{@IT\\@postwita}} & \multicolumn{1}{c}{\makecell[c]{@IT\\@twittiro}} & \multicolumn{1}{c@{}}{\makecell[c]{@TR\\@iwt151}} \\
  \midrule
\textbf{ÚFAL} (ensemble) &@64.17 &@73.6 &@62.7 &@58.6 & 59.1 &@68.3 & 72.2 & 54.7 \\
\textbf{ÚFAL} (single)   & 63.98 &@73.6 & 62.2 & 57.9 & 59.0 &@68.3 & 72.2 & 54.8 \\
HEL-LJU  & 63.72 & 73.5 & 60.6 & 56.2 & 60.3 & 68.1 & @72.3 & @55.0 \\
!MoNoise  &!63.44 &!73.2 &!62.3 &!56.8 &!58.9 &!67.5 &!70.7 &!54.6 \\
!MFR      &!63.31 &!72.9 &!60.3 &!56.7 &!@60.3 &!67.3 &!70.7 &!54.9 \\
TrinkaAI\textsuperscript{*}   & 63.12 & 72.9 & 60.2 & 56.6 & 59.9 & 67.0 & 71.1 & 54.2 \\
maet        & 63.09 & 72.8 & 59.4 & 56.6 & 59.8 & 67.4 & 71.1 & 54.5 \\
team        & 63.03 & 72.8 & 59.4 & 56.6 & 59.8 & 67.2 & 70.9 & 54.5 \\
thunderml\textsuperscript{*}  & 63.02 & 72.7 & 59.6 & 56.7 & 59.2 & 67.3 & 71.3 & 54.2 \\
learnML     & 62.88 & 72.3 & 59.0 & 56.2 & 60.0 & 67.0 & 71.2 & 54.5 \\
CL-MoNoise  & 62.71 & 72.7 & 60.9 & 55.3 & 58.5 & 66.5 & 70.1 & 55.0 \\
BLUE        & 62.53 & 72.6 & 59.6 & 54.2 & 59.8 & 66.7 & 70.0 & 54.8 \\
!LAI         &!62.45 &!72.7 &!59.2 &!53.6 &!60.0 &!66.5 &!70.1 &!@55.0 \\
MaChAmp     & 61.89 & 71.3 & 60.8 & 54.6 & 58.0 & 64.7 & 69.8 & 54.1 \\
  \bottomrule
  \end{tabular}%
  }
  \caption{The results of MultiLexNorm extrinsic evaluation through dependency parsing
  evaluated via label attachment score (LAS). Each team could submit two systems, we show the best of the two. \textsuperscript{*} denotes late submissions.}
  \label{tbl:extrinsic}  
\end{table*}

\subsection{Intrinsic Evaluation}

The intrinsic evaluation is performed using the \textit{Error Reduction Rate} (ERR), which is word-level accuracy normalized to the number of replacements in the dataset. Formally, if we denote the system that does not normalize any word as \textit{leave-as-is}, we can define ERR as
$$\mathit{ERR} = \frac{\textit{accuracy}_\mathit{system} - \textit{accuracy}_\mathit{leave\mhyphen as\mhyphen is}}{1.0 - \textit{accuracy}_\mathit{leave\mhyphen as\mhyphen is}}.$$
The final ranking is determined by the ERR macro-averaged over all datasets.

The MultiLexNorm intrinsic evaluation results are provided in Table~\ref{tbl:intrinsic}. Our system achieves the best performance by a wide margin -- the single model achieves 66.2\% macro-averaged ERR, and the model ensemble even a percent point more, 67.3\%. That is 13.7\% higher than the second-best result of 53.6\%. Our ensemble model reaches best results on all datasets, except for Danish, where our single model is better, and for Spanish, where it is outperformed by other system.

\subsection{Extrinsic Evaluation}

\looseness-1
To evaluate the effect of lexical normalization on downstream applications, MultiLexNorm considers dependency parsing. First, dependency parsing models are trained using the MaChAmp parser~\cite{van-der-goot-etal-2021-massive} on several treebanks from Universal Dependencies 2.8~\cite{zeman-etal-2021-ud2.8}. Treebanks with formal style are used (i.e., not data from social networks), specifically German-GSD, English-EWT, Italian-ISDT and Turkish-IMST. Then, MultiLexNorm participant systems are used to normalize 7 social-media treebanks, which are then parsed using the described parsing models and evaluated using the label attachment score (LAS) metric. For details, please see the MultiLexNorm shared task overview paper.

The results of the MultiLexNorm extrinsic evaluation are presented in Table~\ref{tbl:extrinsic}. Our system also achieves the best performance in the overall macro-average and in 4 out of the 7 treebanks. Generally, the LAS score differences are much smaller than the ERR intrinsic evaluation metric, but the rankings in both evaluations show a lot of similarity. A notable difference is the MFR baseline, which performs remarkably well in the extrinsic evaluation.

\subsection{Ablation Study}
\label{sec:ablations}

\begin{table*}[t!]
  \centering
  \catcode`@ = 13\def@{\bfseries}
  \setlength{\tabcolsep}{4pt}
  \resizebox{\textwidth}{!}{%
  \begin{tabular}{@{}lrccccccccc@{}}\toprule
    \multirow{6}{*}{@Treebank} &
      @Foundation & mT5 & ByT5  & ByT5  & ByT5  & ByT5  & ByT5  & ByT5  & ByT5  & ByT5\\
      & @Pre-training & \NO & \NO & \YES & \YES & \YES & \YES & \YES & \YES & \YES \\
      & @Fine-tuning & base & base & \NO & base & mixed & best & best & best & best \\
      & @Training Data & train & train & train & train & train & train & trn+dev & trn+dev & trn+dev \\
      & @Beam size & 1 & 1 & 1 & 1 & 1 & 1 & 1 & 16 & 16 \\
      & @Ensemble & \NO & \NO & \NO & \NO & \NO & \NO & \NO & \NO & 4 \\
  \midrule
    @Average & & 33.62 & 59.23 & 31.28 & 64.88 & 63.52 & 64.77 & 66.21 & 66.21 & @67.30 \\ 
\midrule
Danish                        &  & 31.65  & 67.41 & 49.37         & 65.82 & 67.72 & 65.82 &@70.25 & @70.25 & 68.67 \\
German                        &  & 42.91  & 59.35 & 49.10         & 63.40 & 62.50 & 63.40 & 65.77 & 65.65 & @66.22 \\
English                       &  & 61.27  & 70.40 & 40.15         & 73.28 & 72.68 & 73.28 & 73.88 & 73.80 & @75.60 \\
Spanish                       &  & ~-0.21 & 43.87 & ~~9.15        & 57.59 & 56.96 & 56.96 & 55.93 & 55.93 & @59.25 \\
Croatian                      &  & 38.11  & 55.15 & 43.31         & 63.18 & 63.03 & 63.03 & 67.29 & 67.29 & @67.74 \\
Indonesian-English\kern-3.1em &  & 50.86  & 63.75 & ~-3.95        & 65.12 & 63.92 & 63.92 & 66.15 & 66.15 & @67.18 \\
Italian                       &  & ~-7.92 & 43.56 & \llap{-}12.87 & 46.53 & 35.64 & 46.53 & 42.57 & 42.57 & @47.52 \\
Dutch                         &  & 43.18  & 55.88 & 43.45         & 62.03 & 62.30 & 62.03 & 62.70 & 62.70 & @63.58 \\
Slovenian                     &  & 56.48  & 71.62 & 57.21         & 78.08 & 77.14 & 78.08 & 79.89 & 79.85 & @80.07 \\
Serbian                       &  & 43.29  & 60.99 & 59.22         & 71.10 & 71.83 & 71.83 & 73.42 & 73.55 & @74.59 \\
Turkish                       &  & 11.82  & 59.80 & 21.79         & 66.55 & 65.03 & 66.55 & 68.41 & @68.58 & @68.58 \\
Turkish-German\kern-3em       &  & 31.99  & 58.98 & 19.46         & 65.82 & 63.45 & 65.82 & 68.27 & 68.19 & @68.62 \\
  \bottomrule
  \end{tabular}%
  }
  \caption{The intrinsic evaluation of the ablation experiments -- we consider various foundation models, whether pre-training is performed, what data do we use for fine-tuning, the decoder beam size and finally whether an ensemble of models is used.}
  \label{tbl:ablations}
\end{table*}

To quantify the effect of various hyperparameters of our system, Table~\ref{tbl:ablations} presents intrinsic evaluation of several ablation experiments.

\vspace{0.5em}
\noindent \textbf{The Foundation Model}~~ We compare the \textit{small} variants of mT5 and ByT5 foundation models when only fine-tuning (and not pre-training on synthetic data) is used. In this setting, the ByT5 model reaches substantially better results (59.2\% average ERR compared to 33.6\%). We therefore did not experiment further with the mT5 model.\footnote{Preliminary experiments on synthetic pre-training of English and Italian also demonstrated considerably worse results of mT5 compared to ByT5.}

\vspace{0.5em}
\noindent\textbf{Pre-training and Fine-tuning Phases}~~ The pre-training phase considerably improves results, reaching 64.8\% ERR compared to 59.2\% ERR without pre-training. We also evaluate the model after the pre-training phase only -- the resulting 31.3\% ERR is quite low, worse than the MFR baseline and most submitted systems.

\vspace{0.5em}
\noindent\textbf{Fine-tuning Data}~~ First, we consider the effect of fine-tuning purely on the MultiLexNorm training data (\texttt{base} fine-tuning in Table~\ref{tbl:ablations}), mixing in the synthetic data with 1:1 ratio (\texttt{mixed} fine-tuning), or selecting the best option according to the development data (\texttt{best} fine-tuning). The results reveal that, unfortunately, our strategy of choosing the best variant based on development performance is actually worse than the pure \texttt{base} fine-tuning. On the other hand, training also on the development data improves the performance substantially (from 64.8\% to 66.2\% average ERR).

\vspace{0.5em}
\noindent\textbf{Beam-search Decoding}~~ Using beam-search decoding with a beam size 16 has virtually no effect compared to greedy decoding (66.21\% average ERR for both options). We hypothesize there is no difference because we generate each normalization independently, so greedy decoding on the \emph{small} target sequences recovers the optimal solution with very high probability. In practice, it is therefore enough to utilize greedy decoding and avoid the higher runtime requirements of a beam search.

\vspace{0.5em}
\noindent\textbf{Ensembling}~~ Finally, utilizing an ensemble of 4 models improves the performance by a percent point from 66.2\% average ERR to 67.3\%.

\begin{table}[t!]
  \centering
  \catcode`@ = 13\def@{\bfseries}
  \setlength{\tabcolsep}{3.8pt}
  \begin{tabular}{@{}lrrr@{}}\toprule
    @GPU & \makecell[c]{@Batch\\@size} & \makecell[c]{@Words\\@per sec} & @Slowdown \\
  \midrule
GeForce RTX 3090 & 1 & 15.45 &~~5.92$\times$ \\
GeForce RTX 3090 & 2 & 28.05 &~~3.26$\times$ \\
GeForce RTX 3090 & 4 & 42.15 &~~2.17$\times$ \\
GeForce RTX 3090 & 8 & 57.55 &~~1.59$\times$ \\
GeForce RTX 3090 & 16 & 84.19 &~~1.09$\times$ \\
GeForce RTX 3090 & 32 & 90.20 &~~1.01$\times$ \\
GeForce RTX 3090 & 64 & 88.01 &~~1.04$\times$ \\
GeForce RTX 3090 & 128 & 91.42 &~~1$\times$~~~~~ \\
GeForce RTX 3090 & 256 & 87.77 &~~1.04$\times$ \\
GeForce RTX 3090 & 512 & 78.94 &~~1.16$\times$ \\
  \midrule
GeForce RTX 1080~\rlap{Ti} & 128 & 55.69 &~~1.64$\times$ \\
GeForce RTX 2080~\rlap{Ti} & 128 & 58.19 &~~1.57$\times$ \\
Quadro P5000 & 128 & 43.88 &~~2.08$\times$ \\
Quadro RTX 5000 & 128 & 66.81 &~~1.37$\times$ \\
  \bottomrule
  \end{tabular}
  \caption{Inference speed of greedy decoding, measured on all English evaluation data (56$\,$999 words in total) as words per second for various GPUs and batch size.}
  \label{tbl:speed}
\end{table}

\subsection{Inference Speed}

To inspect the runtime performance of our model, we measure the inference speed of a single model using greedy decoding on all English evaluation data (56$\,$999 words in total). The results for various batch sizes and GPUs are listed in Table~\ref{tbl:speed}. Overall, with a batch size of 128 our model processes 43-91 words per second, depending on a GPU used. For comparison, the MoNoise system is reported to normalize 29-62 words/sec without a GPU, based on candidate filtering \citep[Table~3]{van-der-goot-2019-monoise}.

\section{Conclusion}

We presented the winning system of the \MultiLexNorm shared task, which is based on a ByT5 foundation model. In intrinsic evaluation, the system performance is superior by a very wide margin, and our system also delivers the best performance in extrinsic evaluation. We release both the source code at
{\small\url{https://github.com/ufal/multilexnorm2021}}
and the fine-tuned models at {\small\url{https://huggingface.co/ufal}}.

In future work, we would like to change the model architecture to encode input sentences only once, either by decoding the whole sentence, or by separating all input words with different sentinel tokens and then decoding individual words by initializing the decoder with the corresponding sentinel tokens.

\section*{Acknowledgements}

The research described herein has been supported by the Ministry of Education, Youths and Sports of the Czech Republic, under the project LINDAT/CLARIAH-CZ (LM2018101).

\bibliography{anthology,custom}
\bibliographystyle{acl_natbib}




\end{document}